\documentclass[12pt]{article}
\usepackage[margin =1in]{geometry}
\usepackage{amssymb,amsthm}
\usepackage{amsmath}
\usepackage[round]{natbib}

\numberwithin{equation}{section}

\newcommand{\inclu}[0] {\ar@{^{(}->}}

\newcommand{\EE}{\mathbb{E}}

\theoremstyle{remark}

\usepackage{mathtools}

\usepackage{pgfplots}
\usepackage{color}
\usepackage{algorithm}
\usepackage{enumitem}
\usepackage{algpseudocode}
\usepackage{caption}
\usepackage{subcaption}

\newcommand{\Var}{\mathrm{Var}}

\usepackage{hyperref} %
\begin{document}

    \title{What is the objective of reasoning with reinforcement learning?}

	
	\author{Damek Davis\thanks{Department of Statistics and Data Science, The Wharton School, University of Pennsylvania;
\url{www.damekdavis.com}. Work supported by NSF DMS award 2047637.} \and Benjamin Recht\thanks{Department of Electrical Engineering and Computer Sciences, University of California, Berkeley
\url{https://people.eecs.berkeley.edu/~brecht/}.}}	
	\maketitle

\begin{abstract}
We show that several popular algorithms for reinforcement learning in large language models with binary rewards can be viewed as stochastic gradient ascent on a monotone transform of the probability of a correct answer given a prompt. In particular, the transformation associated with rejection sampling algorithms is the logarithm and that associated with the GRPO algorithm is the arcsine of the square root.
\end{abstract}

\section{Introduction} Though traditionally associated with sophisticated tree search and approximate dynamic programming, reinforcement learning takes on a unique character in the post-training of large language models. As a tool for adjusting models to align with human preferences or gain proficiency in certain test-taking domains, reinforcement learning algorithms tend to all work like the following meta-algorithm:
\begin{enumerate}
	\item Sample a bunch of prompts from a corpus.
	\item Have the model generate a set of responses to the prompts.
    \item Have some external source, be it crowd workers or a verification engine, label the answers as good or bad.
	\item Fine-tune the model based on the triples of prompts, responses, and labels.
\end{enumerate}
We will refer to this underspecified procedure as Meta-Algorithm 1.

Many algorithms have been proposed that run some sort of variant of the meta-algorithm (see references [8-621]). In this note, we argue that they can all be interpreted as stochastic gradient methods on closely related objective functions. Namely, they aim to find a model that maximizes a monotonically increasing function of the probability of getting a correct answer conditioned on the prompt. 

This serves as a unifying, simple way to look at reinforcement learning applied to large language models and to compare different algorithms. Williams' REINFORCE algorithm and other policy gradient algorithms have enough degrees of freedom that understanding what they do when applied to particular optimization problems is not always transparent. When applied to LLMs, different implementations of Meta-algorithm 1 can all be viewed as maximizing closely related objective functions.

Moreover, Meta-algorithm 1 highlights that the base model must already perform nontrivially in order for fine tuning on correct answers alone to be profitable.  If neither the base model nor the engineer can generate correct answers, then the fine-tuning algorithms cannot make progress. No sophisticated implementation of policy gradient can circumvent this constraint. 

In what follows, after recalling some facts from supervised and reinforcement learning, we present a precise form of Meta-Algorithm 1 in Section~\ref{sec:general}, which aims to maximize the scaled probability of correct responses to prompts via stochastic gradient ascent. We then proceed to show how popular post-training algorithms like REINFORCE, rejection sampling fine-tuning, and GRPO are all special cases of our proposed stochastic gradient method.

\nocite{kingma2014adam}

\section{Background: Supervised Learning and Williams' REINFORCE algorithm.}

It's first worth reminding ourselves of what a supervised learning update looks like for probabilistic modeling. Say our goal is to fit a conditional distribution $p_\theta(y|x)$ to a set of $n$ example input-output pairs $(x_i,y_i)$. A popular approach to this problem maximizes the global objective function
\begin{equation}\label{eq:log-loss}
    \sum_{i=1}^n \log p_{\theta}(y_i| x_i)\,.
\end{equation}
One can motivate this as seeking the maximum likelihood $\theta$ given the observed input-output data. A stochastic gradient of this loss for example $i$ is thus
$$
    \nabla_\theta \log p_{\theta}(y_i|x_i)
$$
We note this because this quantity will appear throughout this work. Most of reinforcement learning when applied to LLMs looks like supervised learning, with the catch that the outputs $y_i$ are generated by the model being optimized.

Let's see how reinforcement learning updates relate to these supervised learning updates. To get there, we review Williams' REINFORCE algorithm \citep{williams1992reinforce} in full generality. REINFORCE is based on the following straightforward observation. Let $p_\theta(z)$ be a probability distribution in $z$. Define the function
$$
    \Phi(\theta) = \mathbb{E}_{z \sim p_{\theta}(\cdot)}[F(z)]\,.
$$
Then for $z$ sampled from $p_\theta$
\begin{equation}\label{eq:basic-williams}
    F(z) \nabla_\theta \log p_{\theta}(z)
\end{equation}
is an unbiased estimate of $\nabla_\theta \Phi(\theta)$.   The REINFORCE algorithm is the stochastic gradient method where the gradient estimate is the concatenation of the unbiased partial derivative estimates of~\eqref{eq:basic-williams}.

The key to deriving this expression is the trivial equality from calculus
$$
    \nabla_\theta \log p_\theta(z) = \frac{\nabla_\theta p_\theta(z)}{ p_\theta(z)} \,.
$$
Hence, by multiplying and dividing gradients of probability distributions, we can derive stochastic gradients that are proportional to the gradients of the logarithms of densities. We will make use of such analyses throughout.

The REINFORCE algorithm is general, and the distribution $p_{\theta}(\cdot)$ does not need to be a conditional distribution like it is in supervised learning. However, reinforcement learning applied to problems of the form
$$
\text{maximize}~~ \sum_{i=1}^n \mathbb{E}_{y \sim p_{\theta}(\cdot|x_i)}[R(y,x_i)]\,.
$$
will have updates that look like supervised learning when the rewards are binary valued. We will explore several examples of this in the next sections.

\section{Reinforcement Learning for Reasoning}\label{sec:general}
Consider the simplest RL problem for large language models. We have a base model $\pi_{\theta}$ from which we can generate \textit{answers}, $y$, to \textit{questions}, $x$. Here $x$ functions as the prompt and $y$, the response, is a random sample from a probability distribution $\pi_{\theta}(\cdot \mid x)$, where $\theta$ denotes the parameters of the probabilistic model. We assume that we have some computational means of determining when an answer $y$ is a ``correct'' response to the question $x$. For each question $x$, we denote the set of correct answers by $C(x)$.

The goal is to fine-tune the probabilistic model $\pi_{\theta}$ so that it frequently correctly answers questions from a corpus of questions $Q$. In what follows, we show how several different versions of Meta-Algorithm 1 can be interpreted as performing stochastic gradient ascent on the objective
\begin{align}\label{eq:monotone_rescaling}
	J_h (\theta) := \mathbb{E}_{x \sim Q} \left[ h\left(   \sum_{y \in C(x)} \pi_{\theta}(y \mid x) \right)\right] \,,
\end{align}
for some monotonically increasing function $h$. The term
\begin{equation}\label{eq:q-theta-def}
   p_\theta(C|x) := \sum_{y \in C(x)} \pi_{\theta}(y \mid x)
\end{equation}
is the probability a sample from $\pi_{\theta}(\cdot \mid x)$ is labeled as correct.

In practice, Meta-Algorithm 1 is implemented through the following more concrete instantiation, which we call Algorithm 1, because we refuse to come up with a cutesy, impenetrable four letter acronym.
\begin{enumerate}
	\item Select a question $x$ from the corpus $Q$.
	\item Sample $M$ answers from the current model. 
    \item Based on the evaluation of each response $y_i$, compute a per-sample weight $Z_i$.
	\item Fine-tune the model with a supervised learning update
	\begin{align}
		\theta \leftarrow \theta + \eta 
        \frac{1}{M} \sum_{i=1}^M Z_i \nabla_\theta \log \pi_{\theta}(y_i \mid x)\,.
	\end{align}
\end{enumerate}

The weights $Z_i$ are often called \emph{advantages} in the RL literature. Typically, algorithm designers motivate the choice of $Z_i$ by appealing to variance reduction. Instead, we show that typical choices of $Z_i$ in fact induce different cost functions $h_M$ such that 
\begin{align}\label{eq:the_expectation}
\EE_{y_{1:M}}\left[\frac{1}{M} \sum_{i=1}^M Z_i \nabla_\theta \log \pi_{\theta}(y_i \mid x)\right] := \nabla_\theta h_M(p_\theta(C \mid x)) 
\end{align}
This means that our choice of advantage determines the objective $J_{h}$ in~\eqref{eq:monotone_rescaling} that we optimize. For example, Williams' version of REINFORCE, affectionately dubbed ``vanilla'' REINFORCE, uses advantage $Z_i = 1_{y_i \in C(x)}$ which clearly leads to objective $h(t)=t$. That is, it attempts to maximize the probability of correct answers, averaged over the questions.

More interestingly, we will show that using rejection sampling to find correct answers corresponds to a function close to $h(t)=\log(t)$. In addition, the popular algorithm GRPO algorithm corresponds to an objective close to $h(t) = \arcsin(\sqrt{t})$.
We also ask the converse question: given a function $h$, how should we choose the weights $Z_i$ so that the induced $h_M$ is close to $h$. It turns out the answer is related to the Bernstein polynomial expansion of $h$; we discuss this in Section~\ref{sec:conclusion}.

\begin{figure}[H]
    \centering
\includegraphics[width=0.45\textwidth]{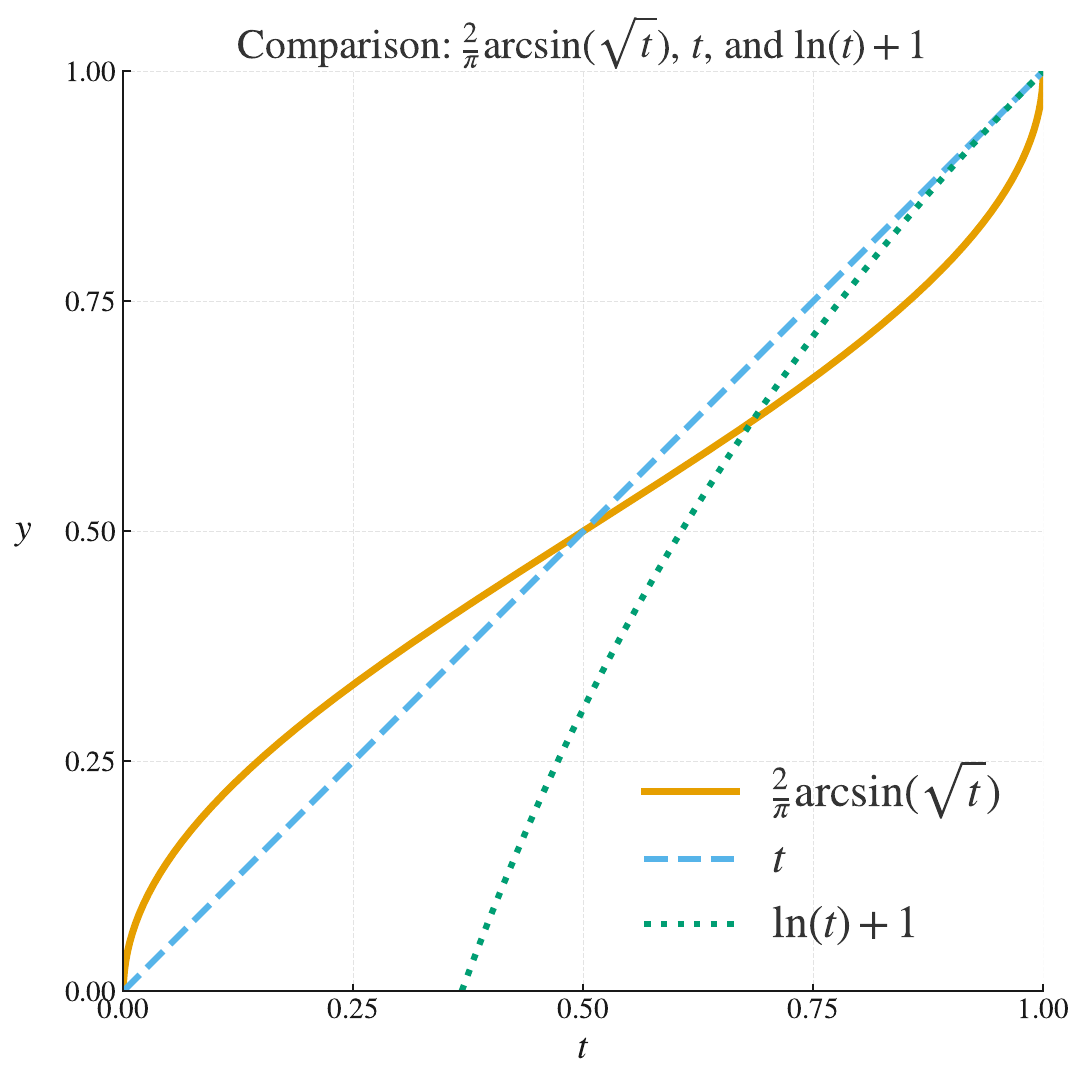}
    \caption{GRPO loss vs. REINFORCE loss vs. log loss}
    \label{eq:figureloss}
\end{figure}

\section{Which functions do the reweightings induce?}\label{sec:reweightings}

We will consider a specific family of weights $Z_i$ which are conditionally linear in $R_i$. Specifically, define rewards $R_i = 1_{y_i \in C(x)}$ and the leave-one-out total rewards $S_i = \sum_{j \neq i} R_i$. The weights we consider in this work are all of the form: 
\begin{align}\label{eq:advantageformula}
Z_i = (1 - R_i)a_{S_i}  +  R_ib_{S_i}
\end{align}
where $s \mapsto a_{s}$ and $s \mapsto b_s$ are arbitrary (measurable) functions of $s$. We claim that that such weights always induce a transform of the form.
\begin{align}\label{eq:costm}
h_M(t) = \frac{1}{M}\sum_{s=0}^{M-1}(b_s - a_s)I_{t}(s+1, M-s),
\end{align}
where $t \mapsto I_{t}(s+1, M-s)$ is the regularized incomplete beta function.  

To prove this, note that each term $Z_i\nabla_\theta\log\pi_\theta(y_i\mid x)$ has the same distribution. Therefore, we must only show
$$
\EE_{y_{i} \sim \pi_{\theta}(\cdot \mid x)}\left[Z_i \nabla_\theta \log \pi_{\theta}(y_i \mid x)\right] = \nabla_\theta h_M(p_\theta(C \mid x)) = h_M'(p_\theta(C \mid x)) \nabla_\theta p_{\theta}(C \mid x).
$$
We can argue this by conditioning on $S_i = s$. Indeed, fix $s$, define a scalar $\phi_s := b_s - a_s$, and define a function $g_s(y)$ which is equal to $b_s$ when $y \in C(x)$ and equal to $a_s$ when $y \notin C(x)$. Then conditioned on $S_i = s$, we have $g_s(y) = Z_i$. Moreover, since $y_i$ is independent of $S_i$, 
$$
\EE\left[Z_i\nabla_\theta\log\pi_\theta(y_i\mid x) \mid S_i=s\right]
=\sum_y g_s(y)\nabla_\theta\pi_\theta(y\mid x)
=\phi_s\sum_{y\in C(x)}\nabla_\theta\pi_\theta(y\mid x)
=\phi_s\nabla_\theta p_\theta(C \mid x),
$$
Averaging over $S_i\sim\mathrm{Bin}(M-1,p)$, we find that 
$$\EE[Z_i\nabla_\theta\log\pi_\theta(y_i\mid x)]=\kappa(p)\nabla_\theta p_\theta(C \mid x) \,,
$$
where 
$$
\kappa(t):=\sum_{s=0}^{M-1}\phi_s\binom{M-1}{s}t^s(1-t)^{M-1-s}.$$
Since $\frac{d}{dt}I_t(s+1,M-s)=M\binom{M-1}{s}t^s(1-t)^{M-1-s}$, we have $h_M'(t)=\kappa(t)$, as desired.

In the next two sections, we consider the weights for rejection sampling and GRPO.

\section{Rejection sampling}\label{sec:rejection}

\begin{figure}[H]
    \centering
\includegraphics[width=0.45\textwidth]{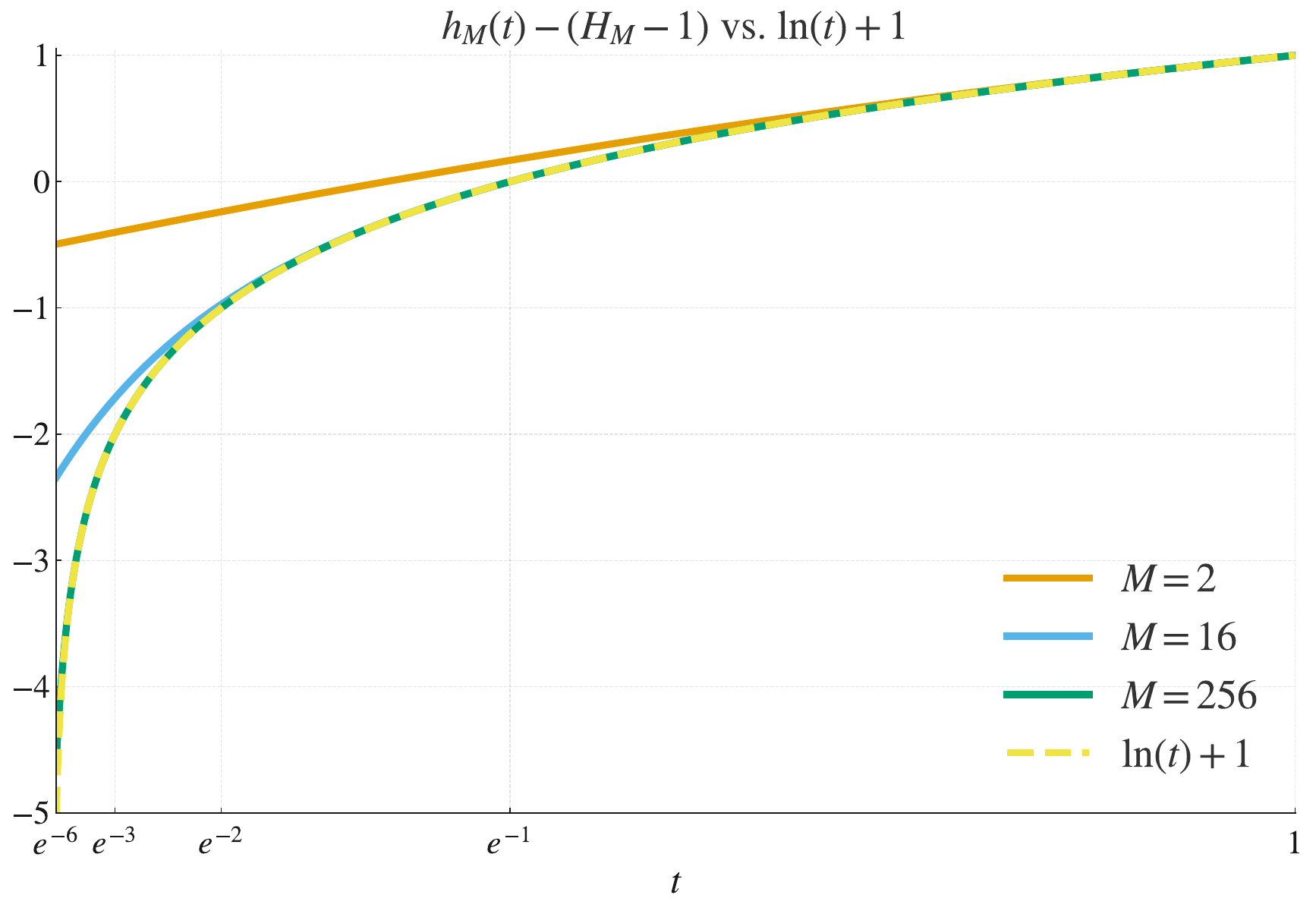}
    \caption{Function $h_M$ induced from rejection sampling.}
    \label{fig:log}
\end{figure}

Consider the special case of Algorithm 1 (see e.g.~\cite{xiong2025minimalist}): Sample $M$ answers from the model, let $V$ denote the set of correct answers. Then use the gradient estimator: 
$$
\frac{1}{|V|}\sum_{y \in V} \nabla_\theta \log \pi_\theta(y \mid x).
$$
If you receive no correct answers, skip the step. 
One can see that the choice of advantage here is simply: 
$
Z_i = R_i M/(S_i + 1) .
$
and thus, the induced cost function is: 
$$
h_M(t) = \sum_{s=0}^{M-1}\frac{1}{s+1}I_{t}(s+1, M-s),
$$
Note that this function is close to $\log(t)$, and gets closer as $M$ approaches infinity. In fact, one can show that 
$$
h_M(t) = \log(t) + H_M + \sum_{r=M+1}^\infty \frac{(1-t)^r}{r},
$$
where $H_M$ is the $M$th harmonic number; see Figure~\ref{fig:log}. 

One might wonder whether it's possible to actually target the objective $h(t) = \log(t)$. This can be achieved by rejection sampling, which falls slightly outside of Algorithm 1. Indeed, consider the following algorithm:
\begin{enumerate}
	\item Select a question from the corpus $Q$.
	\item For a chosen integer, $B$, sample answers from the model until you observe $B$ correct responses. Let $V$ denote this set of correct answers.
    \item Fine-tune the model with a supervised learning update
	$$
		\theta \leftarrow \theta + \eta \frac{1}{B}  \sum_{y\in V} \nabla_\theta \log \pi_\theta( y| x)\,.
	$$
\end{enumerate}
A similar algorithm using rejection sampling was proposed for post-training by \citet{xiong2025minimalist}. The algorithm as written here is in fact stochastic gradient ascent with an unbiased estimate of the gradient of $J_{\log}$. We use an analysis that mimics the log trick used to derive Williams' REINFORCE algorithm. Note that
\begin{align*}
	\nabla_\theta \log \sum_{y \in C(x)} \pi_\theta(y| x)  &=  \frac{\sum_{y \in C(x)}  \nabla_\theta \pi_\theta(y | x)}{ \sum_{y \in C(x)} \pi_\theta(y | x)}\\
	& = \frac{\sum_{y \in C(x)} \pi_\theta(y | x)  \nabla_\theta \log \pi_\theta(y | x)}{ \sum_{y \in C(x)} \pi_\theta(y | x) }\\
	& = \mathbb{E}[  \nabla_\theta \log \pi_\theta(y | x) | y \in C(x), x]\,.
\end{align*}
Hence, sampling $y$ using rejection sampling yields an unbiased stochastic gradient of the cost $J_{\log}$.

Note that this even gives a new way to implement supervised learning, albeit one that is not very practical. For a given example $(x_i, y_i)$, you can sample from $y$ from $p_\theta(\cdot|x_i)$ until you find one with $y=y_i$. At this point, update the model in the direction of the gradient of $\log p_\theta(y|x_i)$. This algorithm maximizes the standard log-loss objective~\eqref{eq:log-loss}. 

However, in the reinforcement learning context, the global objective $J_{\log}$ does not have a natural interpretation as a maximum likelihood estimator unless there is a single correct answer in $C(x)$. The objective $J_{\log}$ is more analogous to the multilabel problem in supervised learning where many labels can be counted as correct for a particular example data-point. As we discuss in the conclusion, it's not clear whether there is a rigorous justification for using the logarithmic scaling in reinforcement learning contexts.

\section{GRPO}
The GRPO algorithm~\citep{shao2024deepseekmath} is another renormalization of the sampled gradients. 
GRPO normalizes by the square root of the variance of the received rewards. This amplifies the impact of each update. Indeed, after sampling $y_1, \ldots, y_M \sim \pi_{\theta}(\cdot  \mid x)$, one forms the stochastic gradient estimator:
\begin{align}
 \frac{1}{M}  \sum_{i=1}^M \left(\frac{R_i - \frac{1}{M}\sum_{i=1}^MR_i}{\sqrt{\Var(\{R_i\}_i)} + \varepsilon}\right)\nabla_\theta \log \pi_\theta( y_i| x)\, \qquad \text{for some $\varepsilon > 0$.}
\end{align}

Which loss does this correspond to?  
Let us first note that conditioned on $S_i =s$, the weighting $Z_i$ is indeed linear in the indicator $R_i$
\begin{align*}
Z_i ^:= \frac{R_i - \frac{1}{M}\sum_{i=1}^MR_i}{\sqrt{\Var(\{R_i\}_i)} + \varepsilon} 
= (1-R_i)\frac{ - q_s}{\sqrt{ q_s(1-q_s)} + \varepsilon} + R_i\frac{1 - q_{s+1}}{\sqrt{ q_{s+1}(1-q_{s+1})} + \varepsilon}
\end{align*}
where $q_s = s/M$. 
From this expression, the objective induced by $Z_i$ is 
$$
J_{M, \varepsilon}(\theta) = \EE_{x \sim Q}\left[h_{M,\varepsilon}\left(\sum_{y \in C(x)} \pi_{\theta}(y \mid x)\right)\right],
$$
where $h_{M, \varepsilon}$ is defined via~\eqref{eq:costm}:
\begin{align*}
h_{M,\varepsilon}(t):=\frac{1}{M}\sum_{s=0}^{M-1} \left(\frac{1 - q_{s+1}}{\sqrt{ q_{s+1}(1-q_{s+1})} + \varepsilon}
+\frac{q_s}{\sqrt{ q_s(1-q_s)} + \varepsilon}\right)\,I_t(s+1,M-s);
\end{align*}
This function is admittedly opaque, but one can get a sense of what it's targeting by considering an idealized setting where $\varepsilon = 0$ and we replace the sample mean and variance by their population values.

\begin{figure}[H]
    \centering
\includegraphics[width=0.75\textwidth]{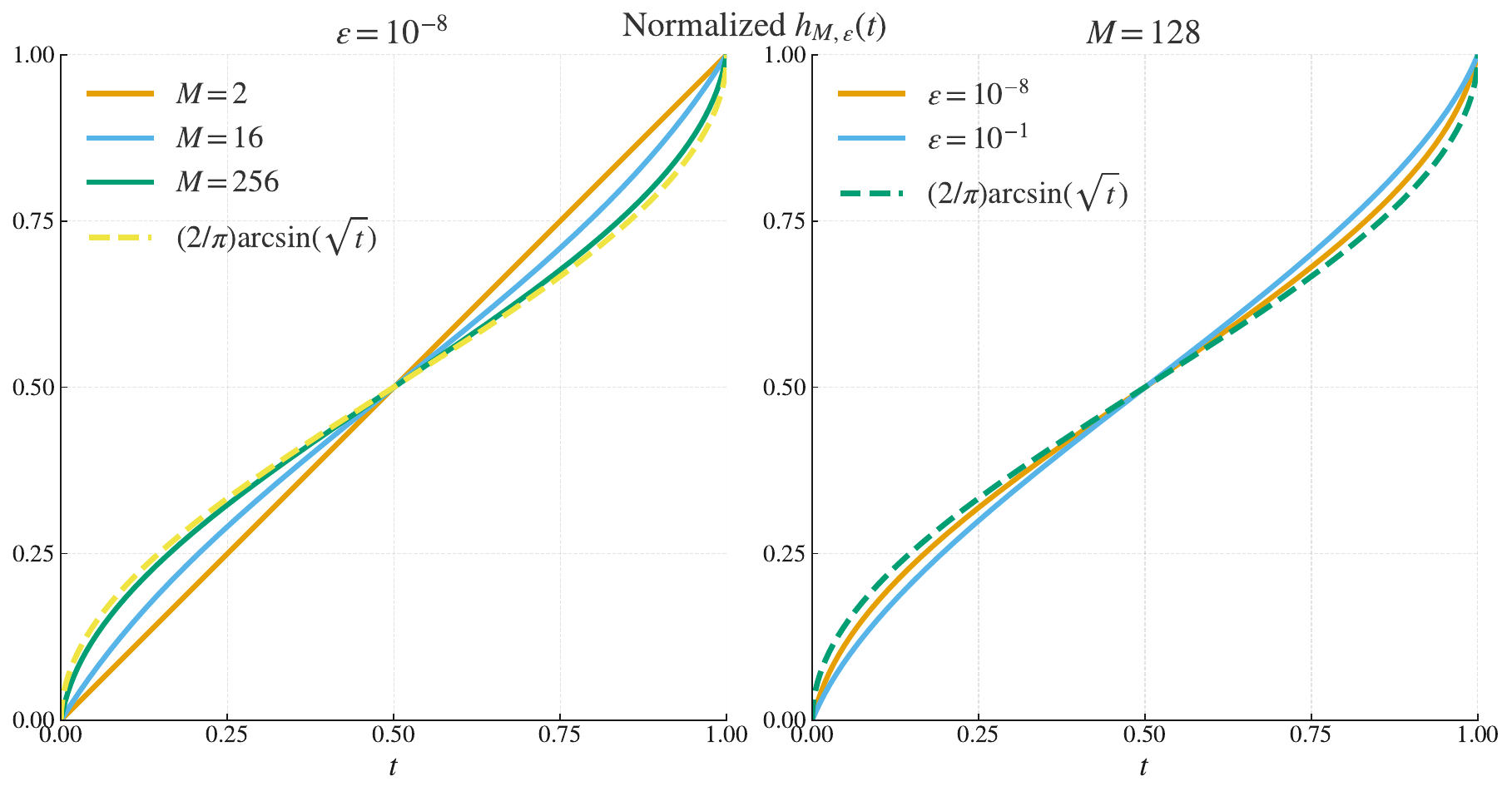}
    \caption{The effect of $M$ and $\varepsilon$ on the function $h_{M, \varepsilon}/h_{M, \varepsilon}(1)$.}
    \label{fig:grpoloss}
\end{figure}

Indeed, let us consider the function $h(t)= 2\arcsin \sqrt{t}$ and differentiate: 
\begin{align*}
\nabla_{\theta} h(p_\theta(C \mid x)) &= \frac{\nabla_\theta p_\theta(C \mid x)}{\sqrt{p_\theta(C\mid x)(1 - p_{\theta}(C \mid x))}}\\
&= \dfrac{\EE_{y \sim \pi_{\theta}(\cdot \mid x)}[R(y, x)\nabla_{\theta}\log\pi_{\theta}(y \mid x)]}{\sqrt{\Var_{y\sim \pi_{\theta}(\cdot \mid x)}(R(y, x))}}\\
&=\EE_{y \sim \pi_{\theta}(y \mid x)}\left[\left(\dfrac{R(y, x) - \EE_{y \sim \pi_{\theta}(\cdot \mid x)}[R(y,x)]}{\sqrt{\Var_{y\sim \pi_{\theta}(\cdot \mid x)}(R(y, x))}}\right)\nabla_{\theta}\log\pi_{\theta}(y \mid x)\right],
\end{align*}
where the third equality uses the fact that the expected value of the log derivative under $\pi_{\theta}(\cdot \mid x)$ is zero. Figure~\ref{fig:grpoloss} plots the behavior of $h_{M, \varepsilon}/h_{M, \varepsilon}(1)$. Reassuringly, we see that for large $M$ and small $\varepsilon$, the function tends to $h(t) = \frac{2}{\pi}\arcsin \sqrt{t}$. As $\varepsilon$ increases, $h_{M, \varepsilon}$ behaves more like $h(t) = t$. 

So again, like the rejection sampling algorithm from Section~\ref{sec:rejection}, the GRPO algorithm is optimizing a monotone rescaling of the probability of achieving a correct answer. Why is this a good scaling function? As we discuss in the conclusion, we're not sure.

\section{What do we learn?}\label{sec:conclusion}

This note only aims to provide language model researchers more clarity and flexibility in designing fine-tuning methods, so we provide no concrete recommendations.
In fact, a variety of other sampling and reweighting methods can be derived using the approach of this note.
For example, replacing the standard deviation by the variance in GRPO yields a function close to the log odds rescaling: $h(t) = \log(t/(1-t))$. Other rescalings based on normalizing by the pdf of the beta distribution have also been considered, e.g., in~\cite{xiao2025bnpo} and their associated loss functions which approximate the corresponding cdfs of the beta distribution can be deduced from~\ref{sec:reweightings}.

What if maximizing log-odds is what you want? Or what if there's some other scaling function $h$ you like? In that case, we can provide you a recipe for constructing a gradient estimator. 
The process is straightforward: the derivative of $h_M$ in~\eqref{eq:costm} is a Bernstein polynomial, which provides a basis in which to approximate any continuous function: 
\begin{align}\label{eq:h_derivatives}
h_M'(t)\;=\;\sum_{s=0}^{M-1}(b_s - a_s)\binom{M-1}{s}t^s(1-t)^{M-1-s}
\end{align}
Then one simply needs to choose $a_s$ and $b_s$ appropriately to best approximate $h'$ by $h_M'$. For example, if $h$ is a polynomial of degree $\leq M$, one may choose these coefficients so that $h_M' = h'$ and hence $h_M = h + \text{const}$. If $h$ is an arbitrary smooth function, one can instead choose coefficients such that $b_s - a_s = h'\left(s/(M-1)\right)$. With this choice of coefficients, $h_M$ converges uniformly to $h + \text{const}$ on $[0,1]$. 
One can even prove a rate of convergence if e.g., $h$ is $C^p$ for $p \geq 3$. In this case the derivative $h_M'$ converges in the $\sup$ norm at a rate $1/M$, and the same result holds for $h_M$ by the fundamental theorem of calculus~\citep{adell2022asymptotic}.

We do not see a reason to prefer any $h$ more than the identity.
For a corpus consisting of a single question, each choice of $h(t)$ induces an identical problem, since $h$ is monotone. 
Unfortunately, it is hard to guess the practical effects of monotone rescaling of the objective function in this manner when there are multiple questions in the corpus. 
In fact, if we allow for an expressive enough representation of $\pi_\theta$ and globally maximize $J_h$, the choice of $h$ is superfluous (though it may affect optimization dynamics): For any monotone scaling, the optimal choice is to place a total probability mass of 1 on $C(x)$  and 0 everywhere else. This is analogous to how hinge and logistic loss on linearly separable data are minimized at the same halfspaces.

Thus, arguing whether GRPO or REINFORCE is best is like arguing whether log loss is better than hinge loss for classification problems. There as well, Logistic regression and support vector machines differ only in a monotonic rescaling of errors in the objective function. Statistically, both return comparable classifiers with sufficient data (See, for example, \citet{bartlett2006convexity}). 
But neither loss has magical properties, and the best loss varies by task.
In the same vein, the best rescaling for RL finetuning will be context dependent, since all such algorithms are chasing closely related objectives.

\bibliography{bibliography}

\begin{thebibliography}{7}
\providecommand{\natexlab}[1]{#1}
\providecommand{\url}[1]{\texttt{#1}}
\expandafter\ifx\csname urlstyle\endcsname\relax
  \providecommand{\doi}[1]{doi: #1}\else
  \providecommand{\doi}{doi: \begingroup \urlstyle{rm}\Url}\fi

\bibitem[Adell and C{\'a}rdenas-Morales(2022)]{adell2022asymptotic}
J.~A. Adell and D.~C{\'a}rdenas-Morales.
\newblock Asymptotic and non-asymptotic results in the approximation by
  bernstein polynomials.
\newblock \emph{Results in Mathematics}, 77\penalty0 (4):\penalty0 166, 2022.

\bibitem[Bartlett et~al.(2006)Bartlett, Jordan, and
  McAuliffe]{bartlett2006convexity}
P.~L. Bartlett, M.~I. Jordan, and J.~D. McAuliffe.
\newblock Convexity, classification, and risk bounds.
\newblock \emph{Journal of the American Statistical Association}, 101\penalty0
  (473):\penalty0 138--156, 2006.

\bibitem[Kingma and Ba(2014)]{kingma2014adam}
D.~P. Kingma and J.~Ba.
\newblock Adam: A method for stochastic optimization.
\newblock \emph{arXiv preprint arXiv:1412.6980}, 2014.

\bibitem[Shao et~al.(2024)]{shao2024deepseekmath}
Z.~Shao et~al.
\newblock Deepseekmath: Pushing the limits of mathematical reasoning in open
  language models.
\newblock \emph{arXiv:2402.03300}, 2024.

\bibitem[Williams(1992)]{williams1992reinforce}
R.~J. Williams.
\newblock Simple statistical gradient-following algorithms for connectionist
  reinforcement learning.
\newblock \emph{Machine Learning}, 8:\penalty0 229--256, 1992.
\newblock \doi{10.1007/BF00992696}.

\bibitem[Xiao et~al.(2025)Xiao, Zhang, and Cao]{xiao2025bnpo}
C.~Xiao, M.~Zhang, and Y.~Cao.
\newblock Bnpo: Beta normalization policy optimization.
\newblock \emph{arXiv preprint arXiv:2506.02864}, 2025.

\bibitem[Xiong et~al.(2025)Xiong, Yao, Xu, Pang, Wang, Sahoo, Li, Jiang, Zhang,
  Xiong, et~al.]{xiong2025minimalist}
W.~Xiong, J.~Yao, Y.~Xu, B.~Pang, L.~Wang, D.~Sahoo, J.~Li, N.~Jiang, T.~Zhang,
  C.~Xiong, et~al.
\newblock A minimalist approach to llm reasoning: from rejection sampling to
  reinforce.
\newblock \emph{arXiv preprint arXiv:2504.11343}, 2025.

\end{thebibliography}
\bibliographystyle{abbrvnat}

\end{document}